\documentclass[runningheads]{llncs}
\usepackage{graphicx}
\usepackage{multirow}
\usepackage{pifont}
\usepackage[table,xcdraw]{xcolor}
\usepackage{subcaption}
\usepackage{enumitem}
\usepackage{amsmath}
\usepackage{amssymb}
\usepackage{booktabs}
\usepackage[misc]{ifsym}
\usepackage[pagebackref,breaklinks,colorlinks,bookmarks]{hyperref}

\begin{document}
%
\title{Rotation Augmented Distillation for Exemplar-Free Class Incremental Learning with Detailed Analysis}
%
%
\author{Xiuwei Chen\inst{1} \and
Xiaobin Chang\inst{1,2,3}$^{(\textrm{\Letter})}$}


\authorrunning{X. Chen, X. Chang}
%
\institute{School of Artificial Intelligence, Sun Yat-sen University, China \and
Guangdong Key Laboratory of Big Data Analysis and Processing, Guangzhou 510006, P.R.China \and
Key Laboratory of Machine Intelligence and Advanced Computing, Ministry of Education, China \\
\email{chenxw83@mail2.sysu.edu.cn, changxb3@mail.sysu.edu.cn}}
%
\maketitle              
\begin{abstract}

Class incremental learning (CIL) aims to recognize both the old and new classes along the increment tasks.
Deep neural networks in CIL suffer from catastrophic forgetting 
and some approaches rely on saving exemplars from previous tasks, known as the exemplar-based setting, to alleviate this problem.
On the contrary, this paper focuses on the Exemplar-Free setting with no old class sample preserved.
Balancing the plasticity and stability in deep feature learning with only supervision from new classes is more challenging.
Most existing Exemplar-Free CIL methods report the overall performance only and lack further analysis.
In this work, different methods are examined with complementary metrics in greater detail.
Moreover, we propose a simple CIL method, Rotation Augmented Distillation (RAD), which achieves one of the top-tier performances under the Exemplar-Free setting.
Detailed analysis shows our RAD benefits from the superior balance between plasticity and stability.
Finally, more challenging exemplar-free settings with fewer initial classes are undertaken for further demonstrations and comparisons among the state-of-the-art methods.


\end{abstract}
%
%
%


\begin{figure*}[t]
\centering
  \begin{subfigure} {0.4\linewidth}
    \includegraphics[width=1\linewidth]{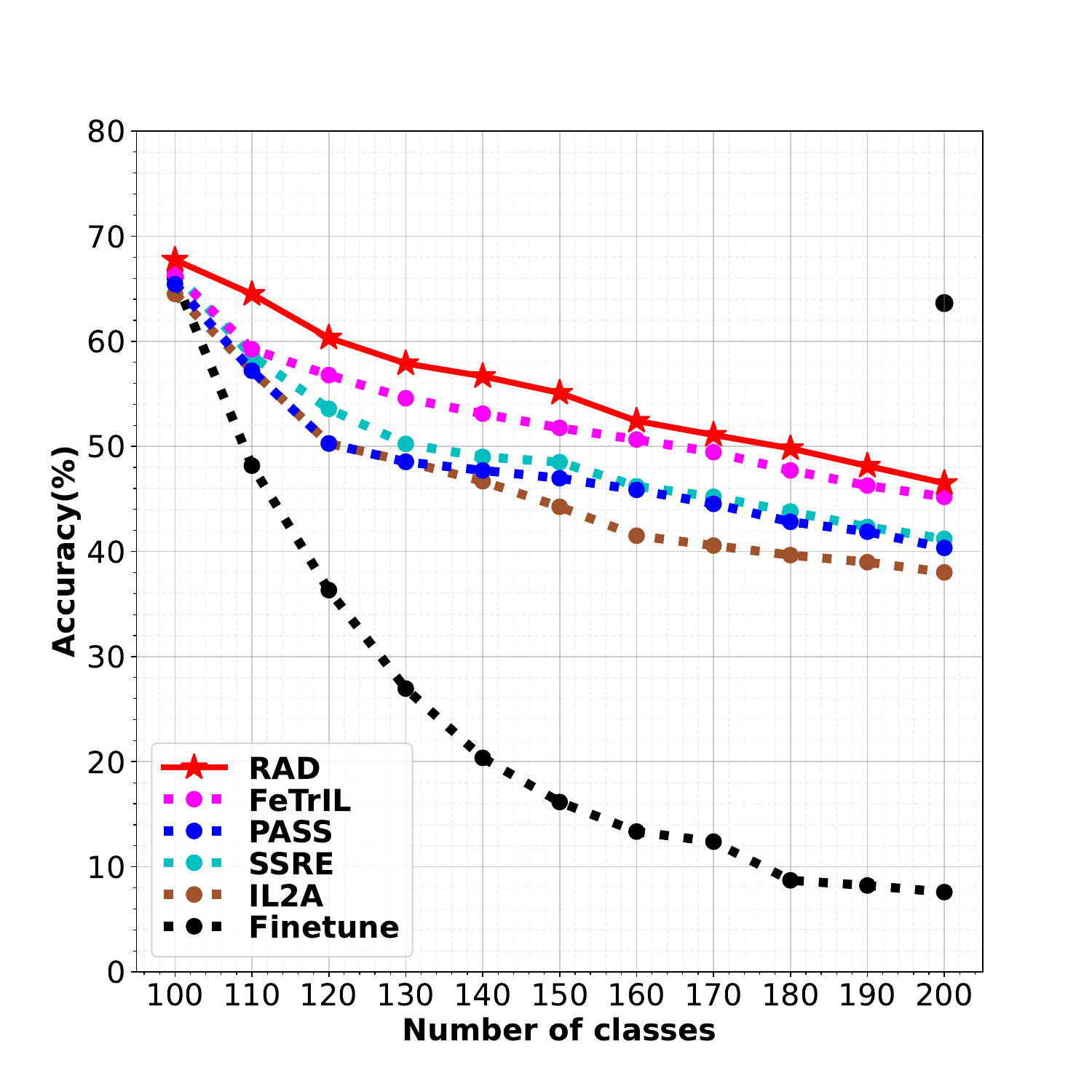}
    \caption{TinyImageNet}
    \label{fig:tiny}
  \end{subfigure}
  \hfill
  \begin{subfigure}{0.4\linewidth}
    \includegraphics[width=1\linewidth]{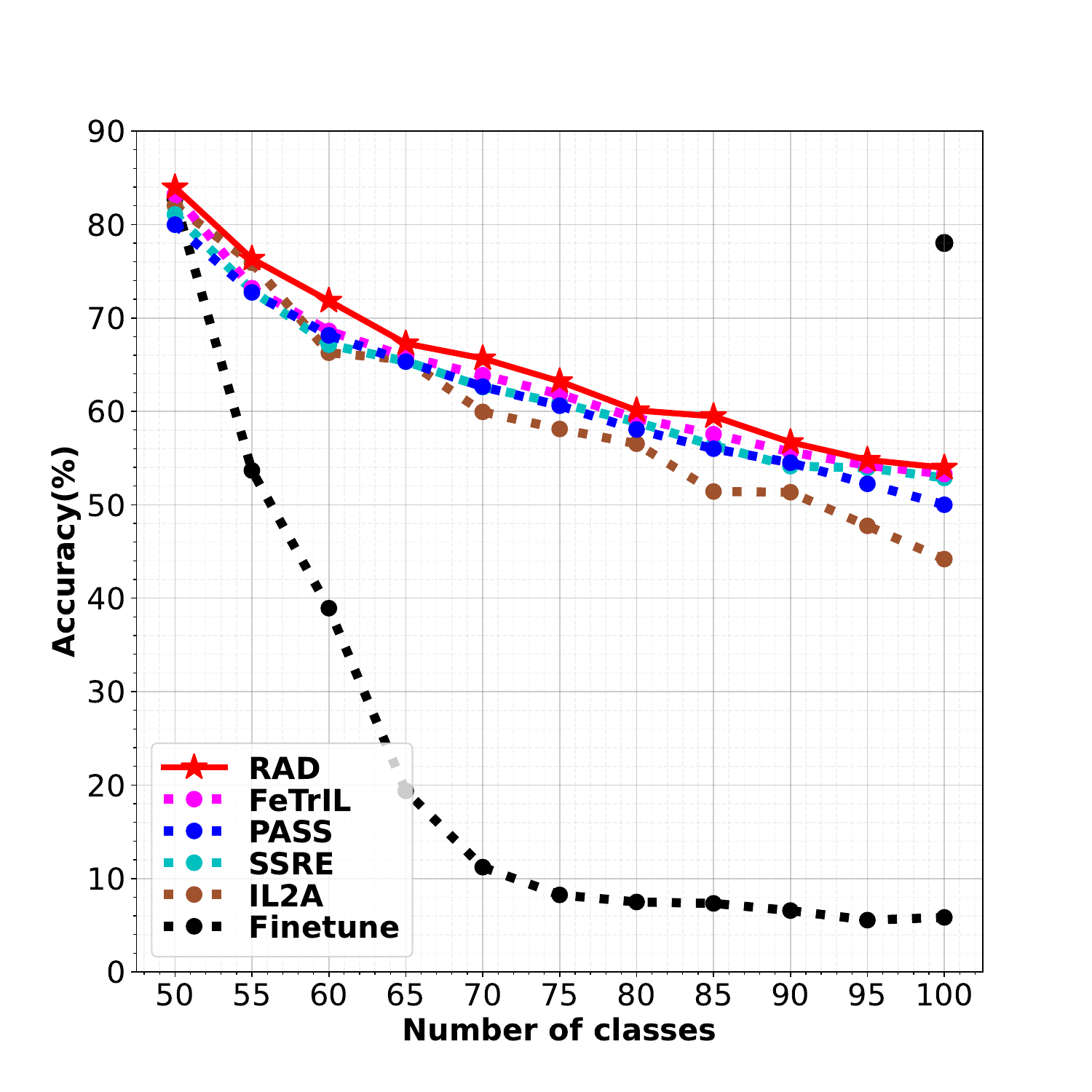}
    \caption{CIFAR100}
    \label{fig:cifar}
  \end{subfigure}
  \hfill
  \caption{
  Incremental Accuracy of TinyImageNet and CIFAR100 with 10 incremental steps.
  The top-1 accuracy (\%) after learning each task is shown.
  Existing SOTA methods achieve similar performance.
  The proposed Rotation Augmented Distillation (RAD) achieves the SOTA performance as well.
  The black dot in the upper right corner indicates the upper bound that the model trained on all the data. The black dotted line indicates the lower bound, a simple finetune method.
  }
  \label{fig:first}
\end{figure*}

\section{Introduction}
\label{sec:introduction}

AI agents, e.g., deep neural networks (DNNs), deployed in the real world face an ever-changing environment, i.e., with new concepts and categories continually emerging~\cite{shaheen2022continual,geiger2012we,li2022coda}.
Therefore, class incremental learning (CIL)~\cite{ercl,masana2020class,icarl} has attracted much attention as it aims to equip deep models with the capacity to continuously handle new categories.
However, learning to discriminate sets of disjoint classes sequentially with deep models is challenging, as they are prone to entirely forgetting the previous knowledge, thus resulting in severe performance degradation of old tasks. It is known as catastrophic forgetting~\cite{catastrophic}.
To maintain the old knowledge, a small number of exemplars from the previous tasks can be stored in the memory buffer, known as the exemplar-based methods~\cite{icarl,ercl,lcinc}.
However, exemplars from previous tasks can be unavailable due to constraints, e.g., user privacy or device limitations.

To this end, this paper focuses on exemplar-free class incremental learning (EFCIL) setting~\cite{ssre,pass,il2a,fetril,sdc}. 
Model distillation~\cite{distill} plays an important role in preserving past knowledge by existing methods~\cite{ssre,pass,il2a,cds}. 
Both the feature extractor and new classifier are learned with the classification loss on the new data and knowledge distillation from the previous model.
The variations of deep features during the new task training are penalized by distillation loss.
Other methods~\cite{fetril,deesil} learn a feature extractor in the initial task only and then fixed it for increment tasks. New classifiers are learned on the extracted features of new tasks while the old class statistics, e.g., prototype vectors and covariance matrices, are also computed and preserved.
The state-of-the-art (SOTA) methods~\cite{ssre,pass,il2a,fetril} follow either the two paradigms mentioned above.
As shown in Fig.~\ref{fig:first}, these SOTA methods report similar overall performance, i.e., average incremental accuracy.
However, further detailed analysis is not conducted within these methods.
In this work, we further incorporate the measures of Forgetting and Intransigence as in~\cite{rwil}.
The performance of SOTA EFCIL methods~\cite{ssre,pass,fetril} under various settings is reproduced for more detailed comparisons and analysis.

As EFCIL is a challenging task, the existing experimental setting is with half of the dataset in the initial task. Therefore, a model trained on such an initial task may be strong enough for the following incremental tasks.
A baseline method, named Feat$^{*}$, is proposed for such demonstration and comparison. Specifically, a deep feature extractor is trained with the initial data and frozen.
NME classifier is then used to discriminate different classes across incremental steps.
However, Feat$^{*}$ achieves the SOTA level performance under the existing setting.
In this work, we propose a simple yet effective method, Rotation Augmented Distillation (RAD), to enable the continuous training of the entire deep model along the increment tasks.
On the one hand, the data augmentation used provides the plasticity by introducing the varied training samples during training.
On the other hand, the knowledge distillation can
achieve the stability by alleviating the forgetting of past knowledge.
To alleviate the bias of the strong initial model, a more challenging setting with much fewer data in the initial task is introduced. Nevertheless, our RAD still achieves one of the best results among the SOTA methods due to its superior balance of stability and plasticity, as revealed by the detailed analysis.

The main contributions of this work are summarized as follows:
\begin{enumerate}[noitemsep]
  \item
  We provide a more detailed analysis of the EFCIL methods, rather than the overall results only in existing works.
  Specifically, the complementary metrics, forgetting and intransigence, are also used to evaluate SOTA methods. Detailed comparisons and analyses are thus enabled.
  \item A simple and intuitive method, Rotation Augmented Distillation (RAD), is designed to better alleviate the plasticity-stability dilemma.
  Its effectiveness is demonstrated by its superior performance under various EFCIL settings.
  \item A new challenging setting is provided to alleviate the bias brought by the strong initial model. Detailed comparisons and analyses are also conducted.
\end{enumerate}

\section{Related Work}

Class incremental learning aims for a well-performing deep learner that can sequentially learn from the streaming data. Its main challenge is Catastrophic Forgetting~\cite{catastrophic} which depicts the deep models prone to entirely forgetting the knowledge learned from previous tasks.
Different strategies have been proposed to handle this issue.
Regularization strategies such as elastic weight consolidation (EWC)~\cite{ocf} use different metrics to identify and penalize the changes of important parameters of the original network when learning a new task.
Rehearsal strategies~\cite{ddrcc,icarl,lsil,wa} are widely adopted as well.
The model is permitted to access data from previous tasks partially by maintaining a relatively small memory, enabling it to directly recall the knowledge from the previous data and mitigate forgetting. 
iCaRL~\cite{icarl} stores a subset of samples per class by selecting the good approximations to class means in the feature space.
However, access to the exemplars of old tasks is not guaranteed and could be limited to data security and privacy constraints~\cite{spl}.

The exemplar-free class incremental learning (EFCIL)~\cite{crla,ssre,pass,il2a,fetril,sdc,deesil} is a challenging setting, where no data sample of old tasks can be directly stored.
Existing EFCIL methods either
use regularization to update the deep model for each incremental step~\cite{ocf,lwm} or adapt distillation to preserve past knowledge by penalizing variations for past classes during model updates~\cite{ssre,pass,il2a}. 
Moreover, the prototypes of old tasks can be exploited in conjunction with distillation to improve overall performance, as shown in~\cite{ssre,pass,il2a}.
Specifically, PASS~\cite{pass} proposes prototype augmentation in the deep feature space to improve the discrimination of classes learned in different increment tasks. 
A prototype selection mechanism is proposed in SSRE~\cite{ssre} to better enhance the discrimination between old and new classes.
Feature generation for past classes is introduced in IL2A~\cite{il2a} by leveraging information about the class distribution. 
However, IL2A is inefficient to scale up for a large number of classes since
a covariance matrix needs to be stored for each class.
Inspired by the transfer learning scheme~\cite{lsimfg}, independent shallow classifiers can be learned based on a fixed feature extractor, as in DeeSIL~\cite{deesil}.
A pseudo-features generator is further exploited in FeTrIL~\cite{fetril} to create representations of past classes to improve classifier learning across tasks.
The proposed Rotation Augmented Distillation (RAD) method is end-to-end trainable on the full model and simply based on a distillation strategy~\cite{distill} and rotation data augmentation~\cite{ssla}. Detailed comparisons show that RAD benefits from the superior balance between stability and plasticity in EFCIL model training.

\section{Methodology}

\subsection{Preliminary}

In class incremental learning, a model is learned from an initial task (0) a sequence of ${T}$ incremental tasks, where each task ${t}$ has a set of $n_t$ different classes ${C_t} = \{ c_{t,1},...,c_{t,n_t}\}$. The classes in different tasks are disjoint, ${C_{i}} \cap {C_j} = \emptyset, i \neq j, i,j \in \{0,...,T\}$.
The training data of task $t$ is denoted as ${D_t}$. ${D_t}$ consists of data tuples $(x,y)$ where $x$ is an image and $y$ is its corresponding ground-truth class label.
A deep classification model $\operatorname{\Phi}$ consists of two modules, the feature extractor $\operatorname{F}$ and the classifier head $\operatorname{h}$. $\operatorname{F}$ is parameterized with $\theta$ and the feature representation of image $x$ is obtained via $\operatorname{F}(x) \in \mathbb{R}^d$. The classifier head $\operatorname{h}$ is parameterized with $\omega$.

Under the exemplar-free class incremental learning (EFCIL) setting, a model can only access ${D_t}$ when training on task $t$. At the initial task, the classification model $\operatorname{\Phi}_0$ is trained under the full supervision of ${D_0}$ and resulting in $\operatorname{F}_0$ and $\operatorname{h}_0$.
At the incremental task $t$, $t \in \{1,...,T\}$, $\operatorname{\Phi}_t$ is partially initialized with $\theta_{t-1}$ and trained with $D_t$. The corresponding feature extractor $\operatorname{F}_t$ and classifier head $\operatorname{h}_t$ are learned.
The overall classifier $\operatorname{H}_t$ is an aggregation of a set of task-specific classifiers $\operatorname{h}_i, i=\{1,...,t\}$, $\operatorname{H}_t = \{ {\operatorname{h}_0},{\operatorname{h}_2}, \cdots ,{\operatorname{h}_{t}} \}$.
During testing, data samples are from all observed classes so far with balanced distributions.

\begin{figure*}[t]
\centering
  \includegraphics[width=0.85\linewidth]{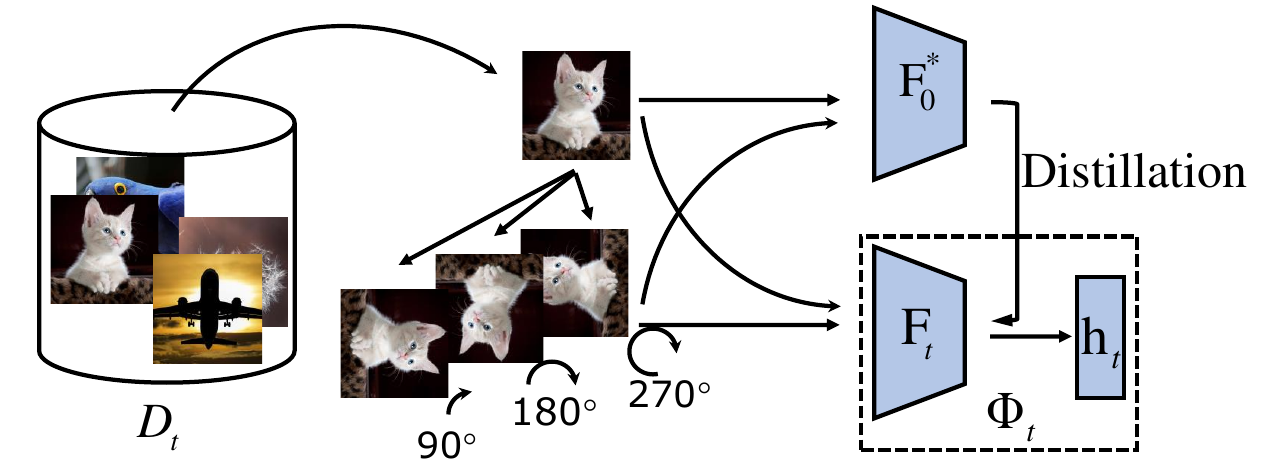}
  \hfill
  \caption{
  Illustrations of the proposed Rotation Augmented Distillation (RAD) method for exemplar-free class incremental learning at task $t$. 
  $^{*}$ indicates the corresponding module is frozen at training.
  }
  \label{fig:main}
\end{figure*}


\subsection{A baseline method: Feat$^{*}$}

The feature extractor learned at the initial task is frozen and denoted as $\operatorname{F}_{0}^{*}$.
The corresponding classifier $\operatorname{h}_{0}$ is abandoned.
$\operatorname{F}_{0}^{*}$ is used as the feature extractor across all tasks and no further training on the deep classifier is needed. This method is denoted as Feat$^{*}$.
$\operatorname{F}_{0}^{*}$ forwards all training samples in $D_t, t \in \{0, ..., T\}$ once for the class-specific mean feature vectors and such prototypes are preserved.
NME classifier can then be applied to the seen prototypes so far during testing.
Different from Feat$^{*}$, a holistic classifier $\operatorname{H}_t$ for the seen classes till task $t$ is trained in FeTrIL based on the pseudo-feature generation. During testing of FeTrIL, $\operatorname{H}_t$ rather than NME classifier is exploited.


\subsection{Rotation Augmented Distillation}

The deep classification model $\operatorname{\Phi}_t$ can be end-to-end learned with two simple techniques, rotation data augmentation and knowledge distillation, as illustrated in Fig.~\ref{fig:main}.
Specifically, for each class, we rotate each training sample $x$ with 90, 180 and 270 degrees~\cite{ssla} and obtained the augmented sample ${{\rm{x'}}}$:
\begin{equation}
    {{\rm{x'}}} = \operatorname{rotate}({x},\delta ),\delta  \in \{ 90,180,270\}.
    \label{eq:rorate}
\end{equation}
Each augmented sample ${{\rm{x'}}}$ subject to a rotated degree is assigned a new label $y'$, extending the original K-class problem to a new $4$K-class problem. The augmented dataset of task $t$ is denoted as $D'_t$.

Both the feature extractor $\operatorname{F}_t$ and the classifier $\operatorname{H}_t$ of $\operatorname{\Phi}_t$ are jointly optimized in RAD.
Based on dataset $D_t$ and the augmented one $D'_t$, the cross-entropy loss is computed,
\begin{equation}
    {\mathcal{L}_c} =\sum\limits_{(x,y) \in {D_t}} {\operatorname{CE}({\operatorname{\Phi}_t}(x),y)} +\sum\limits_{(x',y') \in {D'_t}} {\operatorname{CE}({\operatorname{\Phi}_t}(x'),y')}.
    \label{eq:cross}
\end{equation}
To alleviate the mismatch between the saved old prototypes and the feature extractor, the knowledge distillation~\cite{distill} is employed to regularize learning of the feature extractor. Specifically, we restrain the feature extractor by matching the features of new data extracted by the current model with that of the initial model $\operatorname{F}_{0}^{*}$:
\begin{equation}
    {\mathcal{L}_{distil}} = \sum\limits_{(x,y) \in {D_t}} {\operatorname{KL}({\operatorname{F}_t}({x};{\theta _t})||{\operatorname{F}_{0}^{*}}({x};{\theta _{0}}))} + \sum\limits_{(x',y') \in {D'_t}} {\operatorname{KL}({\operatorname{F}_t}({x'};{\theta _t})||{\operatorname{F}_{0}^{*}}({x'};{\theta _{0}}))}. 
    \label{eq:kd}
\end{equation}
The total loss of RAD comprises two terms, 
\begin{equation}
    {\mathcal{L}_{all}} = \alpha{\mathcal{L}_c} + \beta{\mathcal{L}_{distil}},
    \label{eq:all loss}
\end{equation}
with $\alpha$ and $\beta$ are balancing hyper-parameters. Both of them are set to 1 across all experiments.
The learning objective of RAD becomes $\min_{\theta_t, \omega_t} \mathcal{L}_{all}$.


\section{Experiments}

{ \bf Datasets} 
The exemplar-free class incremental learning (EFCIL) is conducted on three datasets. Two large-scale datasets are TinyImageNet~\cite{tinyimagenet} and ImageNet100~\cite{imagenet}. The medium-size dataset used is CIFAR100~\cite{cifar100}.

\noindent{\bf Protocols}\quad
Two EFCIL settings are followed in our experiments.
(1) Conventional EFCIL setting. The conventional setting usually with half of the data as initial tasks.
TinyImageNet contains images from 200 classes in total and the initial task (0) includes half of the dataset, i.e., images from 100 classes, denoted as \textbf{B100}. The data of the remaining classes are split into $T$ tasks. For example, $T=$\textbf{5} corresponds to 5 incremental steps and each step contains the data of 20 classes. Therefore, this EFCIL setting of TinyImageNet is denoted as {\textbf{B100 5 steps}}. {\textbf{B100 10 steps}} and {\textbf{B100 20 steps}} are the other two EFCIL protocols of TinyImageNet.
Both CIFAR100 and ImageNet100 are with 100 classes.
They have the following three EFCIL protocols: {\textbf{B50 5 steps}}, {\textbf{B50 10 steps}} and {\textbf{B40 20 steps}}.
(2) Challenging EFCIL setting. Training a method with half of the data can result in a strong initial model, which can introduce a non-negligible bias towards the following incremental learning performance. Therefore, a more challenging EFCIL setting is proposed by reducing the data in the initial task to half or even less. Specifically, 
the challenging EFCIL settings of TinyImageNet are {\textbf{B50 5 steps}}, {\textbf{B50 10 steps}} and {\textbf{B50 25 steps}}.


\noindent{\bf Implementation details}\quad
The proposed Rotation Augmented Distillation (RAD) and the reproduced SOTA EFCIL methods~\cite{ssre,pass,fetril} are implemented with PyTorch~\cite{pytorch} and trained and tested on NVIDIA GTX 3080Ti GPU.
Other results are from~\cite{fetril}.
The initial model is trained for 200 epochs, and the learning rate is 0.1 and gradually reduces to zero with a cosine annealing scheduler. For incremental tasks, we train models for 30 epochs, and the initial learning rate is 0.001 with a cosine annealing scheduler.
For a fair comparison, we adopt the default ConvNet backbone: an 18-layer ResNet~\cite{resnet}. 
The reproduced results of SOTA methods are with the same augmentation and assigned to incremental tasks using the same random seed as ours.


\noindent{\bf Evaluation metrics}\quad
Three complementary metrics are used throughout the experiments.
{\bf Overall performance} 
is typically evaluated by average incremental accuracy~\cite{icarl}. After each batch of classes, the evaluation result is the classification accuracy curve. If a single number is preferable, the average of these accuracies is reported as average incremental accuracy.
{\bf Forgetting}~\cite{rwil}
is defined to estimate the forgetting of previous tasks. The forgetting measure for the $t$-th task after the model has been incrementally trained up to task $k$ as:
\begin{equation}
 f_t^k = \mathop {\max }\limits_{l \in \{0, \cdots ,k - 1\}} ({a_{l,t}} - {a_{k,t}}),\forall t < k.
\label{eq:forgetting}
\end{equation}
Note, $a_{m,n}$ is the accuracy of task $n$ after training task $m$. The average forgetting at $k$-th task is then defined as ${\mathbb{F}_k} = \frac{1}{{k }}\sum\nolimits_{i = 0}^{k - 1} {f_i^k} $. Lower $\mathbb{F}_{k}$ implies less forgetting on previous tasks.
{\bf Intransigence}~\cite{rwil}
measures the inability of a model to learn new tasks. We train a reference model with dataset $ \cup _{t = 0}^k{D_t}$ and measure its accuracy on the held-out set of the $k$-th task, denoted as $a_k^-$. The intransigence for the $k$-th task as:
\begin{equation}
 {\mathbb{I}_k} = a_k^- - {a_{k,k}},
\label{eq:intrasigence}
\end{equation}
where $a_{k,k}$ denotes the accuracy on the $k$-th task when trained up to task $k$ in an incremental manner. Note, the lower the $\mathbb{I}_k$ the better the model.

\begin{table}[t]
\caption{
Results of different EFCIL methods under the conventional setting.
The overall performance, average incremental accuracy ($\%$), is reported.
Best results in red, second best in blue. 
}\label{tab:main results}
\centering
\scalebox{0.9}{
\begin{tabular}{cccccccccc}
\hline
                                  & \multicolumn{3}{c}{TinyImageNet}                                                                                                                             & \multicolumn{3}{c}{ImageNet100}                                                                                                                          & \multicolumn{3}{c}{CIFAR100}                                                                                                                                    \\ \cline{2-10} 
\multirow{-2}{*}{Methods}      & 5 steps                                  & 10 steps                                 & 20 steps                                                                 & 5 steps                                  & 10 steps                                 & 20 steps                                                               & 5 steps                                  & 10 steps                                 & 20 steps                                                                    \\ \hline

Finetune & 28.8  & 24.0  & 21.6  & 31.5  & 25.7  & 20.2  & 27.8  & 22.4  & 19.5 \\

EWC~\cite{ocf}                                                                                               & 18.8                                 & 15.8                                 & 12.4                                                                  & -                                    & 20.4                                 & -                            & 24.5                                 & 21.2                                 & 15.9                                              \\
LwF-MC~\cite{icarl}                                                                                          & 29.1                                 & 23.1                                 & 17.4                                                                 & -                                    & 31.2                                 & -                                     & 45.9                                 & 27.4                                 & 20.1                                       \\
DeeSIL~\cite{deesil}                                                                                              & 49.8                                 & 43.9                                 & 34.1                                                                   & 67.9                                 & 60.1                                 & 50.5                      & 60.0                                 & 50.6                                 & 38.1                                                 \\
LUCIR~\cite{lucir}                                                                                          & 41.7                                 & 28.1                                 & 18.9                                                                    & 56.8                                 & 41.4                                 & 28.5                                      & 51.2                                 & 41.1                                 & 25.2                                      \\
MUC~\cite{muc}                                                                                              & 32.6                                 & 26.6                                 & 21.9                                                                    & -                                    & 35.1                                 & -                         & 49.4                                 & 30.2                                 & 21.3                                                      \\
SDC~\cite{sdc}                                                                                              & -                                    & -                                    & -                                    & -                                                                      & 61.2                                 & -                                & 56.8                                 & 57.0                                 & 58.9                                                \\
ABD~\cite{abd}                                                                                                & -                                    & -                                    & -                                    & -                                    & -                                                                       & -                        & 63.8                                 & 62.5                                 & 57.4                                                     \\
IL2A~\cite{il2a}                                                                                           & 47.3                                 & 44.7                                 & 40.0                                                                    & -                                    & -                                    & -                       & 66.0                                 & 60.3                                 & 57.9                                                          \\
PASS~\cite{pass}                                                                                               & 49.6                                 & 47.3                                 & 42.1                                                                    & 64.4                                 & 61.8                                 & 51.3                 & 63.5                                 & 61.8                                 & 58.1                                                          \\
SSRE~\cite{ssre}                                                                                           & 50.4                                 & 48.9                                 & 48.2                                                                    & 65.4                                    & 62.1                                 & 58.8                     & 65.9                                 & 65.0                                 & {\color[HTML]{3166FF} \textbf{61.7}}                                                             \\
FeTrIL~\cite{fetril}                          & {\color[HTML]{3166FF} \textbf{54.8}} & {\color[HTML]{3166FF} \textbf{53.1}} & 52.2                                                                 & {\color[HTML]{3166FF} \textbf{72.2}} & {\color[HTML]{3166FF} \textbf{71.2}} & {\color[HTML]{3166FF} \textbf{67.1}}     & {\color[HTML]{3166FF} \textbf{66.3}} & {\color[HTML]{3166FF} \textbf{65.2}} & 61.5    \\ \hline
Feat$^{*}$                                                                                  & 53.3                                 & 53.0                                 & {\color[HTML]{3166FF} \textbf{52.8}} & 70.1                                 & 69.9                                 & 65.1                                                              & 63.9                                 & 63.6                                 & 59.8          \\
RAD                                                                & {\color[HTML]{FE0000} \textbf{55.9}}                                 & {\color[HTML]{FE0000} \textbf{55.6}}                                 & {\color[HTML]{FE0000} \textbf{55.2}}                                                                & {\color[HTML]{FE0000} \textbf{72.4}}                                 & {\color[HTML]{FE0000} \textbf{71.8}}                                 & {\color[HTML]{FE0000} \textbf{67.4}}                                                                &  {\color[HTML]{FE0000} \textbf{66.5}}                                 & {\color[HTML]{FE0000} \textbf{65.3}}                                 & {\color[HTML]{FE0000} \textbf{61.9}}            \\
                                  \hline
\end{tabular}}
\end{table}

\begin{figure*}[t]
\centering
  \begin{subfigure} {0.32\linewidth}
    \includegraphics[width=1\linewidth]{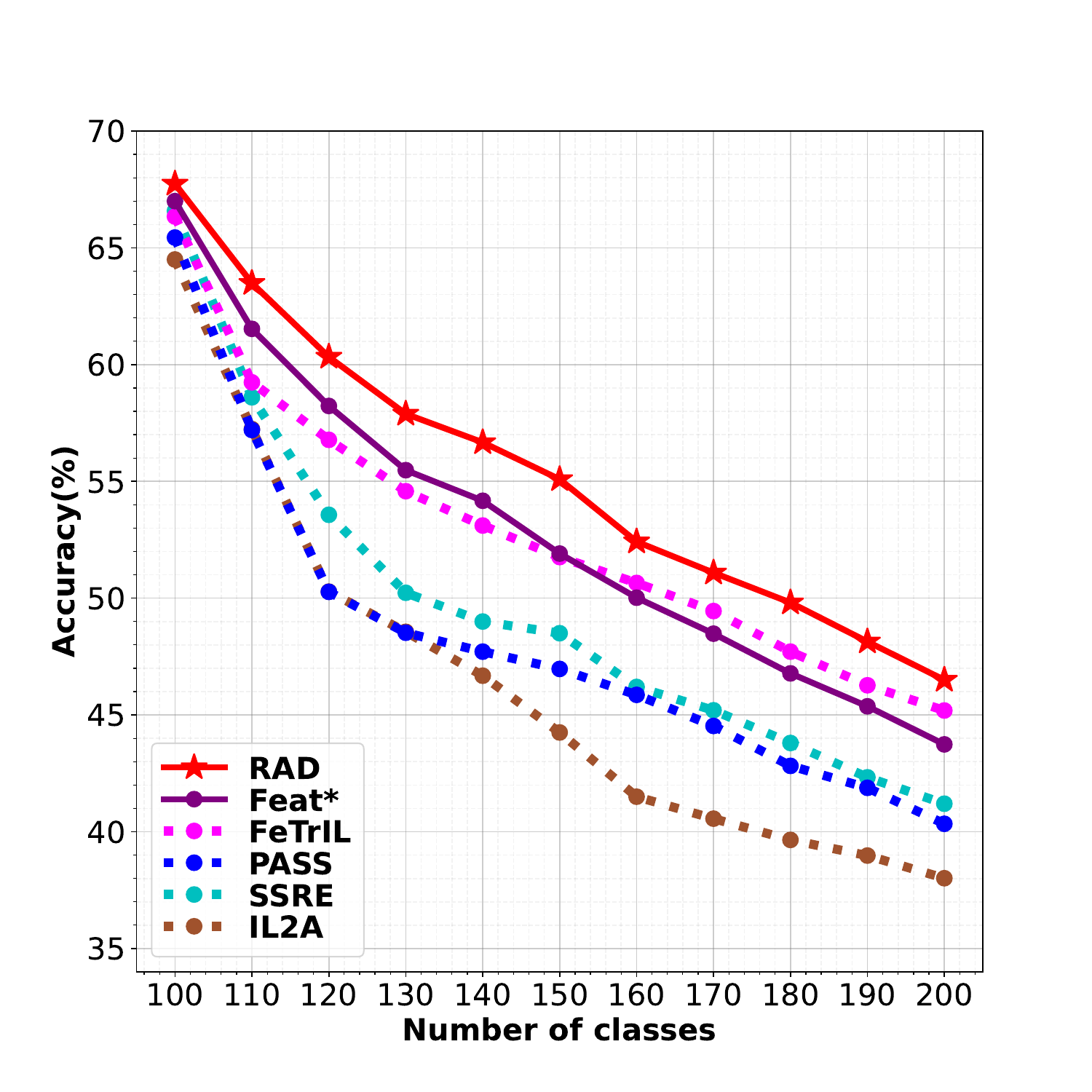}
    \caption{TinyImgNet B100 10}
    \label{fig:tiny b100}
  \end{subfigure}
  \hfill
  \begin{subfigure}{0.32\linewidth}
    \includegraphics[width=1\linewidth]{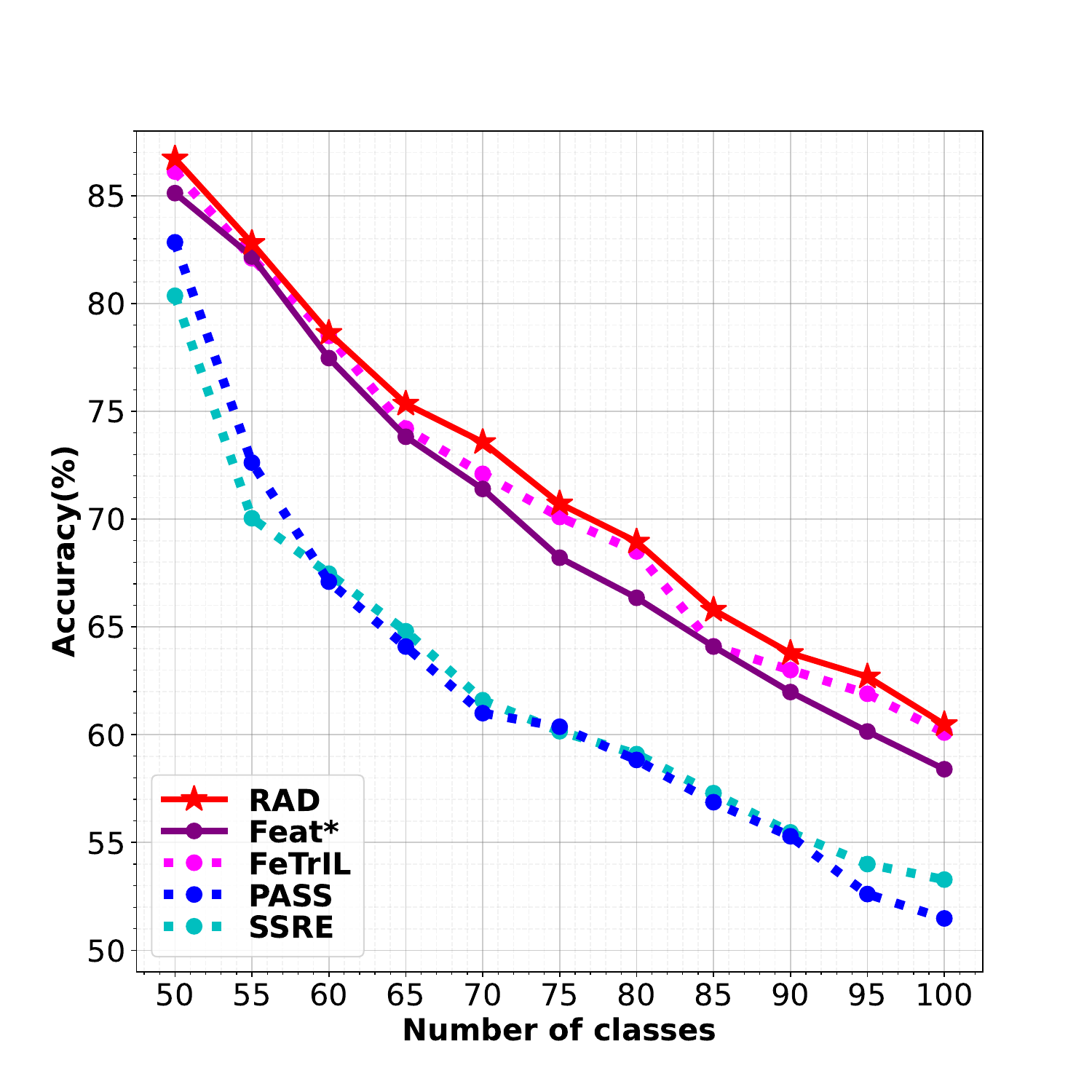}
    \caption{ImageNet100 B50 10}
    \label{fig:cifar b50}
  \end{subfigure}
  \hfill
  \begin{subfigure}{0.32\linewidth}
    \includegraphics[width=1\linewidth]{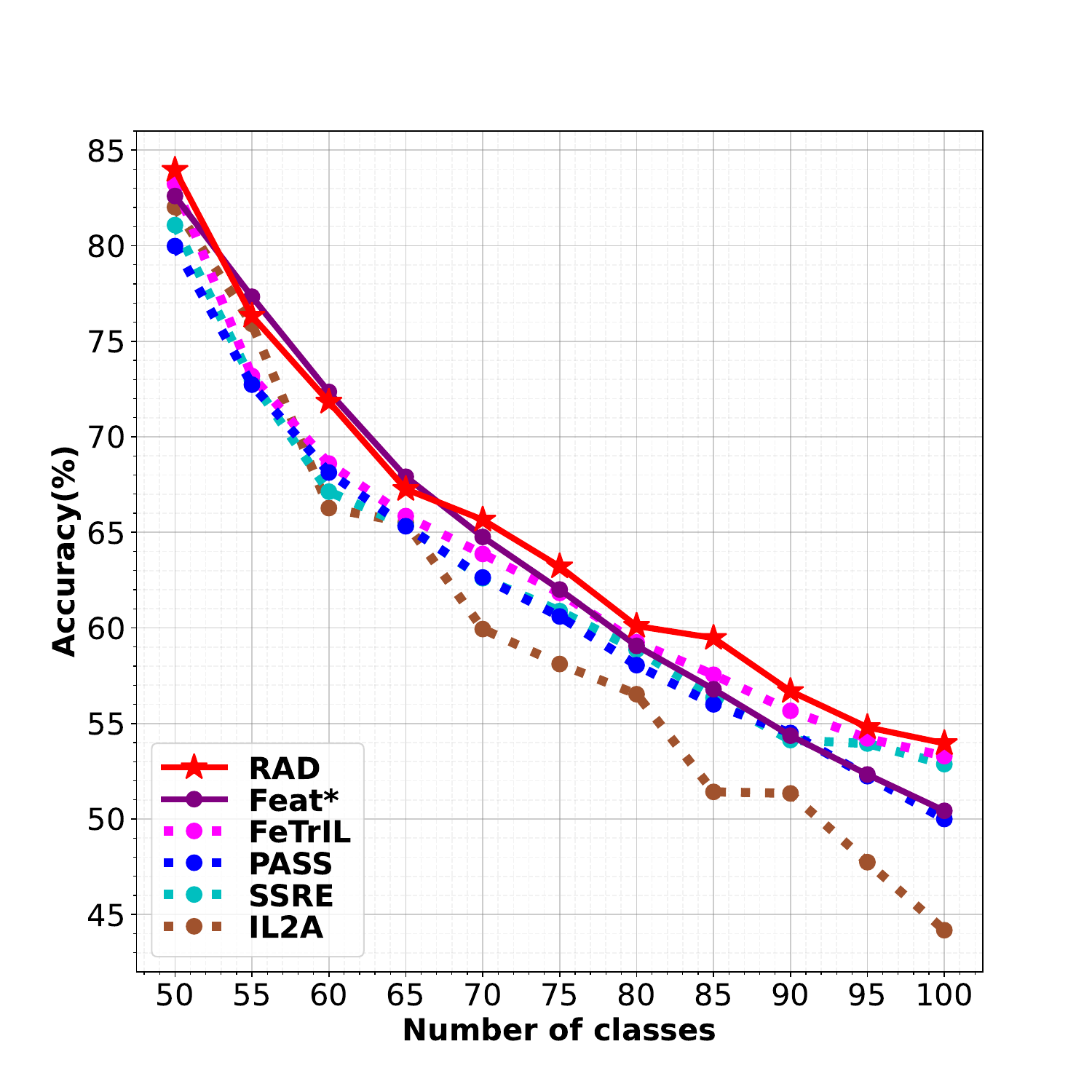}
    \caption{CIFAR100 B50 10}
    \label{fig:tiny B50}
  \end{subfigure}
  \hfill
  \caption{
  Incremental Accuracy Curves of the SOTA methods.
  Each point represents the incremental classification accuracy (\%) after model learning on each task.
  }
  \label{fig:imagenet}
\end{figure*}

\subsection{Detailed Analysis under conventional EFCIL setting}
\label{sec:main results}

The overall results of different methods under conventional EFCIL settings of TinyImageNet, ImageNet100 and CIFAR100 are reported in Tab.~\ref{tab:main results}.
The simple baseline method, Feat$^{*}$, achieves similar results compared to the SOTA methods. For example, the margins between FeTrIL and Feat$^{*}$ are no more than $2\%$. Feat$^{*}$ is observed to be slightly better ($0.6\%$ improvements) than FeTrIL under the TinyImageNet B100 20 steps setting.
Therefore, it demonstrates that a model trained with half of the dataset at the initial task may already equip sufficient classification capability for the subsequent incremental tasks even without further learning. This conclusion is frequently discussed in the previous sections of this paper and inspires the introduction of the challenging EFCIL setting.
The proposed RAD consistently achieves the best results under conventional EFCIL settings of all datasets, as shown in Tab.~\ref{tab:main results}.
For example, RAD is better than FeTrIL on the largest dataset, TinyImageNet.
The more incremental steps the larger improvements.
Such improvement increased from $1.1\%$ under 5 steps to $2.5\%$ under 10 steps and reached $3.0\%$ under 20 incremental steps.
Detailed comparisons among different methods along the incremental learning procedure are also illustrated in Fig.~\ref{fig:imagenet}.
The curve of a superior method is usually above the inferior ones.
For example, the superiority of RAD over FeTrIL can be demonstrated by their curves in Fig.~\ref{fig:imagenet}(a).

\begin{table}[t]
\caption{Results on TinyImageNet and ImageNet100 with intransigence and forgetting metrics. $\mathbb{I}$ represents Intransigence ($\%$), $\mathbb{F}$ represents Average Forgetting ($\%$).
Best results in red, second best in blue. 
}\label{tab:two metrics}
\centering
\begin{tabular}{ccccccc|cccccc}
\hline
\multirow{3}{*}{Methods} & \multicolumn{6}{c|}{TinyImageNet}                                                              & \multicolumn{6}{c}{ImageNet100}                                                          \\ \cline{2-13} 
                         & \multicolumn{2}{c}{5 steps} & \multicolumn{2}{c}{10 steps} & \multicolumn{2}{c|}{20 steps} & \multicolumn{2}{c}{5 steps} & \multicolumn{2}{c}{10 steps} & \multicolumn{2}{c}{20 steps} \\ \cline{2-13} 
                         & $\mathbb{I}$ ($\downarrow$)  & $\mathbb{F}$ ($\downarrow$) & $\mathbb{I}$ ($\downarrow$)  & $\mathbb{F}$ ($\downarrow$)  & $\mathbb{I}$ ($\downarrow$)  & $\mathbb{F}$ ($\downarrow$)  & $\mathbb{I}$ ($\downarrow$)  & $\mathbb{F}$ ($\downarrow$) & $\mathbb{I}$ ($\downarrow$)  & $\mathbb{F}$ ($\downarrow$)  & $\mathbb{I}$ ($\downarrow$)  & $\mathbb{F}$ ($\downarrow$)  \\ \hline
PASS~\cite{pass}                      & 22.5                                 & 15.4                                & 24.3                                 & 20.6                                & 29.6                                 & 25.2                                 & 27.0                                 & 19.3                                & 32.0                                 & 25.7       & 42.4                                 & 31.6       \\
SSRE~\cite{ssre}                      & 21.3                                 & 16.1                                & 23.2                                 & 21.1                                & 25.4                                 & 24.3                                 & 25.7                                 & 24.9                                & 30.2                                 & 30.5       & 34.0                                 & 35.0       \\
FeTrIL~\cite{fetril}                    & {\color[HTML]{3166FF} \textbf{18.9}} & 12.6                                & {\color[HTML]{3166FF} \textbf{19.4}} & 11.8                                & 21.0                                 & 12.9                                 & {\color[HTML]{3166FF} \textbf{21.7}} & 14.7                                & {\color[HTML]{3166FF} \textbf{23.4}} & 15.4       & {\color[HTML]{3166FF} \textbf{27.7}} & 18.1       \\ \hline
Feat$^{*}$                  & 20.9                                 & {\color[HTML]{3166FF} \textbf{8.9}} & 20.9                                 & {\color[HTML]{FE0000} \textbf{9.0}} & {\color[HTML]{3166FF} \textbf{20.9}} & {\color[HTML]{FE0000} \textbf{8.9}}  & 25.1                                 & {\color[HTML]{FE0000} \textbf{8.9}} & 25.1                                 & {\color[HTML]{FE0000} \textbf{9.8}} & 31.9                                 & {\color[HTML]{FE0000} \textbf{10.4}}       \\
RAD                      & {\color[HTML]{FE0000} \textbf{18.0}} & {\color[HTML]{FE0000} \textbf{8.7}} & {\color[HTML]{FE0000} \textbf{18.1}} & {\color[HTML]{3166FF} \textbf{9.6}} & {\color[HTML]{FE0000} \textbf{18.4}} & {\color[HTML]{3166FF} \textbf{10.2}} & {\color[HTML]{FE0000} \textbf{21.5}} & {\color[HTML]{3166FF} \textbf{9.7}} & {\color[HTML]{FE0000} \textbf{23.0}} & {\color[HTML]{3166FF} \textbf{11.4}}       & {\color[HTML]{FE0000} \textbf{27.6}} & {\color[HTML]{3166FF} \textbf{11.0}}          \\
 \hline
\end{tabular}
\end{table}


To provide more detailed analyses on the SOTA EFCIL methods, their performance on another two metrics, forgetting and intransigence, are reported in Tab.~\ref{tab:two metrics}.
The proposed baseline Feat$^{*}$ achieves the lowest $\mathbb{F}$ under most cases. It shows that Feat$^{*}$ suffers the least from forgetting since the deep model is frozen at the initial state and does not fine-tuned on the following tasks.
Moreover, the $\mathbb{I}$ of Feat$^{*}$ still achieves mid-level performance. It further demonstrates the that the initial model trained on half of the dataset can provide sufficient discriminative power for the rest classes, as described above.
FeTrIL achieves the best overall performance among the exisitng SOTA methods, as shown in Tab.~\ref{tab:main results}. Therefore, both $\mathbb{I}$ and $\mathbb{F}$ of FeTrIL are lower than those of its counterpart SOTA methods, as shown in Tab.~\ref{tab:two metrics}.
Similarly, the proposed RAD achieves the best overall performance on various datasets. This is consistent with the results in Tab.~\ref{tab:two metrics}. The $\mathbb{I}$ and $\mathbb{F}$ of RAD are either the best or second best results. 
Specifically, RAD is only slightly worse than Feat$^{*}$ on the $\mathbb{F}$ metric.
This is compensated by the best $\mathbb{I}$ performance of RAD.
Moreover, RAD is consistently better than FeTrIL on both metrics.
Therefore, the superiority of our RAD relies on the more balanced performance between plasticity and stability.

\begin{table}[t]
\caption{
Ablative Study of the proposed RAD. Results of TinyImageNet B100 10 steps and ImageNet100 B50 10 steps are reported.
}
\label{tab:component}
\centering
\begin{tabular}{cc|ccc|ccc}
\hline
\multicolumn{2}{c|}{Components} & \multicolumn{3}{c|}{TinyImageNet} & \multicolumn{3}{c}{ImageNet100} \\ \hline 
        Rotation       & Distillation      & Avg ($\uparrow$)         & $\mathbb{I}$ ($\downarrow$)   & $\mathbb{F}$ ($\downarrow$)         & Avg ($\uparrow$)           & $\mathbb{I}$ ($\downarrow$)  & $\mathbb{F}$ ($\downarrow$)           \\ \hline
                   &        & 24.0         & 57.0        & 76.1          & 25.7   & 77.3          & 90.2          \\
             \checkmark   &    & 11.2   & 60.1      & 74.3        &     13.7      & 78.7   & 87.7       \\
        & \checkmark                 & 52.9         & 21.1     & 11.6   & 70.0          & 24.2    & 26.7      \\
         \checkmark          & \checkmark        & 55.6         & 18.1   & 9.6     & 71.8          & 23.0   & 11.4       \\ \hline
\end{tabular}
\end{table}

\begin{figure}[t]
\centering
  \begin{subfigure}{0.4\linewidth}
    \includegraphics[width=1\linewidth]{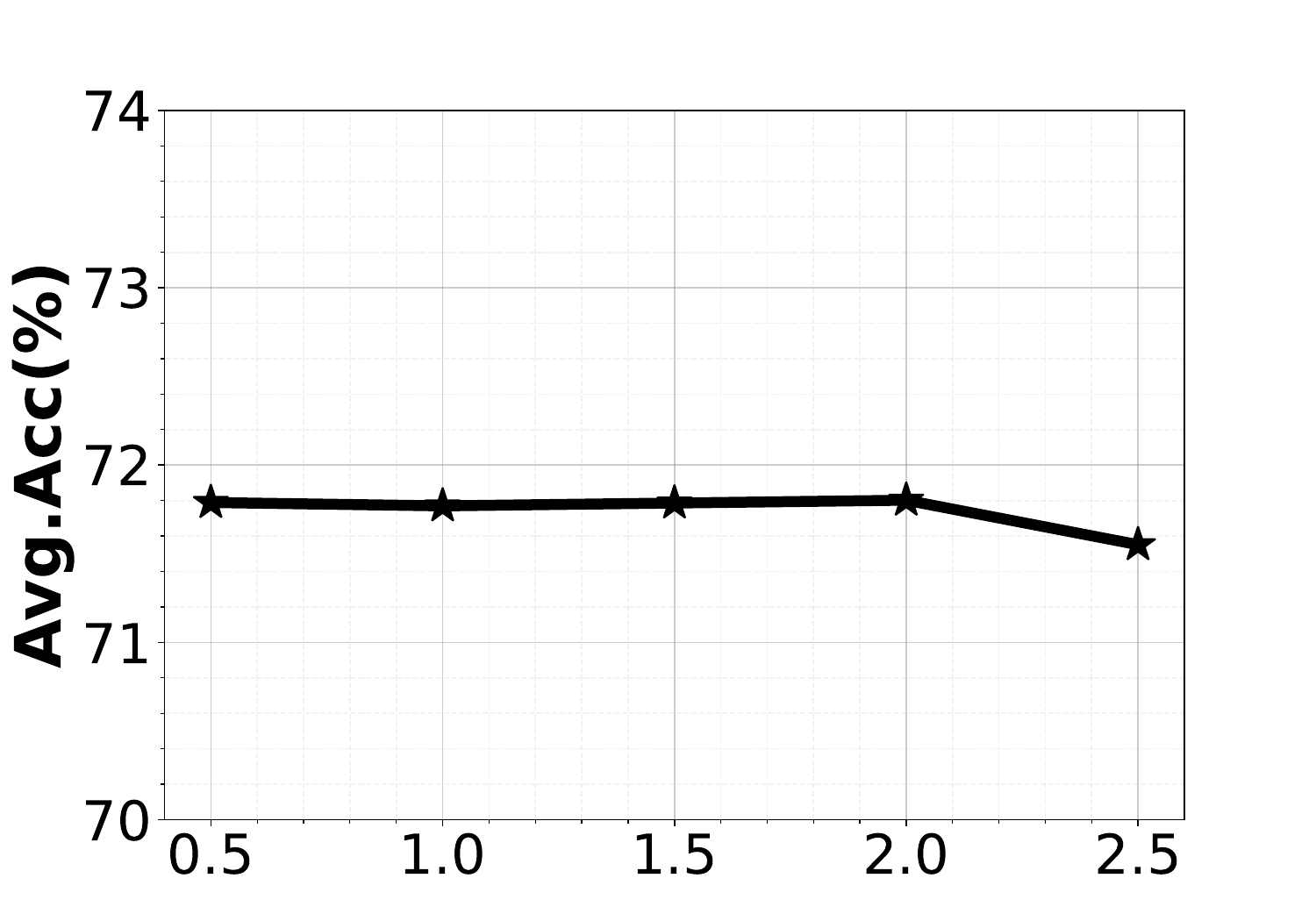}
    \caption{$\alpha$}
    \label{fig:alpha}
  \end{subfigure}
  \hfill
  \begin{subfigure}{0.4\linewidth}
    \includegraphics[width=1\linewidth]{figs/parameter_alpha.pdf}
    \caption{$\beta$}
    \label{fig:beta}
  \end{subfigure}
  \caption{
  Impacts of varying $\alpha$ and $\beta$ on overall results of ImgNet100 B50 10 steps.
  }
  \label{fig:parameter}
\end{figure}

\noindent{\bf Ablation study}\quad
Contributions of different components
in the proposed RAD are illustrated in Tab.~\ref{tab:component}.
{
First, applying data rotation in model finetuning harms the overall performance,
as the augmented images seem to alleviate the forgetting issue (lower $\mathbb{F}$) but make less discriminative new knowledge learned (higher $\mathbb{I}$).
Second, model distillation clearly boosts the overall performance.
Distillation helps the model learning not only to defy forgetting the old knowledge but also to better capture the discriminative information from new tasks, 
as its $\mathbb{F}$ and $\mathbb{I}$ are relatively low.
Finally, RAD combines these two techniques.
Detailed analysis shows that model distillation benefits from rotation data augmentation and then achieves a superior balance between plasticity and stability.
}

\noindent{\bf Impacts of balancing hyper-parameters}\quad
{There are two balancing hyper-parameters, $\alpha$ and $\beta$.  They are set to $1.0$ by default.
As shown in Fig.~\ref{fig:parameter}, Our method is not sensitive to such changes.}

\subsection{Detailed Analysis under challenging EFCIL setting}

The results of challenging EFCIL setting are reported in Tab.~\ref{tab:forgetting on new setting}.
Comparing the overall performance, the average incremental accuracy, of the methods in Tab.~\ref{tab:forgetting on new setting} with those in Tab.~\ref{tab:main results},
all of them experience significant drops, usually more than {$6\%$}, due to much less data being provided in the initial task under the new setting.
Specifically, the baseline Feat$^{*}$ suffers from the worst performance degradation, from $9.2\%$ to $10.1\%$, under different incremental steps.
This is because Feat$^{*}$ is a simple CIL method that relies on the initial task training only.
With less initial data provided, the corresponding model can be much less generalizable on the new data from unseen classes.
Under the challenging EFCIL setting, the proposed RAD still achieves the best overall performance against the existing SOTA methods.
With the $\mathbb{I}$ and $\mathbb{F}$ of different methods compared in Tab.~\ref{tab:forgetting on new setting}, RAD consistently achieve either the best or second best performance on these metrics. It demonstrates that RAD can better alleviate the plasticity and stability dilemma in EFCIL than existing methods.

\begin{table}[t]
\caption{
Results of TinyImageNet under challenging EFCIL setting with average incremental accuracy (Avg.), intransigence ($\mathbb{I}$) and forgetting metrics ($\mathbb{F}$) reported.
Best results in red, second best in blue.
}
\label{tab:forgetting on new setting}
\centering
\begin{tabular}{cccccccccc}
\hline
                          & \multicolumn{3}{c}{5 steps}                                                 & \multicolumn{3}{c}{10 steps}                & \multicolumn{3}{c}{25 steps}                                                                                 \\ \cline{2-10} 
\multirow{-2}{*}{Methods} & Avg. ($\uparrow$) & $\mathbb{I}$ ($\downarrow$)                                    & $\mathbb{F}$ ($\downarrow$)       & Avg. ($\uparrow$)    & $\mathbb{I}$ ($\downarrow$)                                    & $\mathbb{F}$ ($\downarrow$) & Avg. ($\uparrow$) & $\mathbb{I}$ ($\downarrow$)                                    & $\mathbb{F}$ ($\downarrow$)                                                            \\ \hline
PASS~\cite{pass}           & 45.1           & 30.3                                 & 15.5          & 42.5                       & 32.0                                 & 20.2        & 39.7    & 34.6   & 26.5                                      \\
SSRE~\cite{ssre}         & 43.6             & 32.2                                 & 12.2 & 41.1 & 33.7                                 & 14.8   & 40.1       & 33.9   & 21.5                                        \\
FeTrIL~\cite{fetril}         & {\color[HTML]{3166FF} \textbf{47.9}}           & {\color[HTML]{FE0000} \textbf{29.7}} & 13.5    & {\color[HTML]{3166FF} \textbf{46.5}}                             & {\color[HTML]{3166FF} \textbf{30.4}} & 13.4     & {\color[HTML]{3166FF} \textbf{45.1}}      & {\color[HTML]{3166FF} \textbf{30.8}}   & 13.9                                     \\ \hline
Feat$^{*}$           & 44.1       & 34.6                                 & {\color[HTML]{FE0000} \textbf{10.1}} & 43.3 & 34.6                                 & {\color[HTML]{FE0000} \textbf{9.9}}  & 42.7 & 34.6   & {\color[HTML]{FE0000} \textbf{10.1}}    \\
RAD               & {\color[HTML]{FE0000} \textbf{48.6}}       & {\color[HTML]{3166FF} \textbf{30.2}} & {\color[HTML]{3166FF} \textbf{11.6}}        & {\color[HTML]{FE0000} \textbf{47.7}}                         & {\color[HTML]{FE0000} \textbf{30.3}} & {\color[HTML]{3166FF} \textbf{11.5}} & {\color[HTML]{FE0000} \textbf{47.0}} & {\color[HTML]{FE0000} \textbf{30.2}}   & {\color[HTML]{3166FF} \textbf{12.3}}    \\ \hline
\end{tabular}
\end{table}

\section{Conclusion}

We provide a more detailed analysis and comparison of different exemplar-free class incremental learning (EFCIL) methods than many existing works.
Besides the overall performance, i.e., the average incremental accuracy, two complementary metrics, the Forgetting and the Intransigence, are included to measure the EFCIL methods from the perspectives of stability and plasticity.
A simple yet effective EFCIL method, Rotation Augmented Distillation (RAD), has been proposed.
RAD consistently achieves one of the state-of-the-art overall performances under various EFCIL settings.
Based on the detailed analysis, we find that the superiority of RAD comes from the more balanced performance between plasticity and stability.
Moreover, a challenging EFCIL new setting with much fewer data in initial tasks is proposed. It aims to alleviate the bias of a strong initial pre-trained model and stand out the incremental learning performance.

\noindent {\bf Acknowledgement} \quad
This research is supported by the National Science Foundation for Young Scientists of China (No. 62106289).

\end{document}